\documentclass[letter, 11pt, conference]{ieeeconf}      
\IEEEoverridecommandlockouts
\usepackage{cite}
\usepackage{color}
\usepackage{graphicx}
\usepackage{epstopdf}
\usepackage[cmex10]{amsmath}
\def\Real{\mathbb{R}}
\usepackage{bbm}
\usepackage{algorithm,algpseudocode}

\usepackage{amsmath,amssymb}
\usepackage{lipsum}
\usepackage{comment}
\usepackage{array}
\input{mysymbol.sty}
\usepackage{amsthm,bm}
\usepackage{dsfont}
\usepackage{url}
\usepackage[
top    = 1cm,
bottom = 1cm,
left   = 1cm,
right  = 1cm]{geometry}

\usepackage{subfigure}
\usepackage{tikz}
\usepackage{pgfplots}
\usepgfplotslibrary{groupplots}
\usetikzlibrary{external}
\tikzsetexternalprefix{figures/}
\tikzset{external/optimize=false}

\definecolor{mygreen}{rgb}{0.10,0.50,0.10}

\usepackage{ifthen}
\newboolean{showcomments}
\setboolean{showcomments}{true}
\usepackage{todonotes}

\newcommand{\santiago}[1]{  \ifthenelse{\boolean{showcomments}}
{\todo[inline,color=cyan]{Santiago: #1}}{}}

\newcommand{\weiqin}[1]{  \ifthenelse{\boolean{showcomments}}
{\todo[inline,color=orange]{Weiqin: #1}}{}}

\usepackage{fancyhdr}

\theoremstyle{definition}

\author{Weiqin Chen \\
\textit{Department of Electrical, Computer, and Systems Engineering} \\
\textit{Rensselaer Polytechnic Institute}\\
Troy, NY, USA \\
chenw18@rpi.edu}

\renewcommand{\comment}[1]{}
\newcolumntype{S}{>{\centering\arraybackslash} m{.10\linewidth} }
\newcolumntype{T}{>{\centering\arraybackslash} m{.30\linewidth} }

\title{\textbf{Open Problems and Modern Solutions for Deep Reinforcement Learning}}

\begin{document}

\maketitle

\thispagestyle{fancy}
\lhead{}
\chead{} 
\rhead{}
\lfoot{} 
\cfoot{} 
\cfoot{\thepage}
\renewcommand{\headrulewidth}{0pt}
\renewcommand{\footrulewidth}{0pt}

\pagestyle{fancy}
\cfoot{\thepage}

\begin{abstract}
Deep Reinforcement Learning (DRL) has achieved great success in solving complicated decision-making problems. Despite the successes, DRL is frequently criticized for many reasons, e.g., data inefficient, inflexible and intractable reward design. In this paper, we review two publications that investigate the mentioned issues of DRL and propose effective solutions. One designs the reward for human-robot collaboration by combining the manually designed extrinsic reward with a parameterized intrinsic reward function via the deterministic policy gradient, which improves the task performance and guarantees a stronger obstacle avoidance. The other one applies selective attention and particle filters to rapidly and flexibly attend to and select crucial pre-learned features for DRL using approximate inference instead of backpropagation, thereby improving the efficiency and flexibility of DRL. Potential avenues for future work in both domains are discussed in this paper.
\end{abstract}

\section{Introduction}\label{sec_intro}

Deep reinforcement learning (DRL) has gained attraction as one of the closest mechanisms that look anything like artificial general intelligence (AGI)~\cite{goertzel2007artificial}, since it does not require labels as in supervised learning~\cite{cunningham2008supervised}. DRL trains the agent through a trial-and-error way using only reward feedback from the environment, which allows autonomous learning for the agent and substantially reduces human intervention\cite{gu2017deep}.
DRL has emerged as a crucial framework to solve complicated decision-making problems~\cite{sutton2018reinforcement} and has achieved great success in many
challenging and high-dimensional tasks, e.g., playing video games~\cite{mnih2013playing}, mastering Go~\cite{silver2017mastering}, robotic manipulation~\cite{levine2016end} and locomotion~\cite{duan2016benchmarking}, etc. 

Despite the successes, DRL has been criticised for many aspects including difficult reward design, data inefficient, inflexible, stuck in the local optima, etc~\cite{rlblogpost}. These are part of reasons that lead to the difficulty of DRL to be employed in real-world applications e.g., human-robot collaboration (HRC)~\cite{liu2021deep,qureshi2018intrinsically} and autonomous driving~\cite{kiran2021deep}, although many state-of-the-art algorithms like DQN~\cite{mnih2013playing},
DDPG~\cite{lillicrap2015continuous}, TRPO~\cite{schulman2015trust}, PPO~\cite{schulman2017proximal} have been proposed and developed.

Recently, there have been several modern approaches developed that can rescue DRL from the aforementioned issues. For example in terms of reward function design in HRC, one can manually design a fitness function~\cite{singh2009rewards} or an extrinsic reward function~\cite{jaderberg2016reinforcement} regarding the optimization goal based on one's prior knowledge. However, the fitness function is generally rudimentary and lacks guidance for agents' decisions. On the other hand, HRC naturally involves two main objectives of task completion and human's safety, which requires a weight contribution for each objective to trade-off from each other. Under this context, the naive fitness function and extrinsic reward function perform terrible in the HRC problem since one has to manually tune the optimal weight contribution frequently. One approach based on the optimal reward framework~\cite{singh2009rewards, singh2010intrinsically} to deal with the above issue is to introduce an auxiliary parameterized intrinsic reward function as a compensation to the manually designed extrinsic reward and merges them together to generate the final reward~\cite{liu2021deep}, which reduces the manual adjustment of the weight contribution and is discussed in detail in Section~\ref{sec_reward_design}.

Nevertheless, as described in~\cite{rlblogpost}, DRL still suffers from data inefficiency and inflexibility even given an appropriate reward function. To solve these issues, one can apply DRL to the pre-learned representations learned from raw inputs through a task-agnostic way like unsupervised learning~\cite{bengio2012deep}. This helps due to the fact that the dimensionality of the input for DRL has been reduced and the pre-learned features are not tuned to be task-specific. This approach, however, is limited by the number of pre-learned features and suffers from the fact that the previously learned function will be entirely overwritten once the task changes because of backpropagation. Accordingly, instead of employing DRL in an end-to-end way through backpropagation, one might consider having access to some pre-learned representations and then select subsets of the representations by combining the selective attention mechanism~\cite{paneri2017top} with particle filters~\cite{blakeman2022selective}. We describe the details of this setting in Section~\ref{sec_feature_selection}. In Section~\ref{section_discussion} we discuss and provide some future research avenues for both reward function design and feature selection. Section~\ref{sec_conclusions} completes this paper with concluding remarks.

\section{Reward Function Design for HRC}\label{sec_reward_design}
As discussed in Section~\ref{sec_intro}, many industrial applications of DRL such as HRC and autonomous driving suffer from the tricky problem of reward function design. This section reviews a solution to the problem in HRC domain~\cite{liu2021deep}.
\subsection{Human-robot Collaboration Problem}
\begin{figure}[!ht]
	\begin{minipage}[t]{0.49\linewidth}
		\centering
		\includegraphics[width=1.5in]{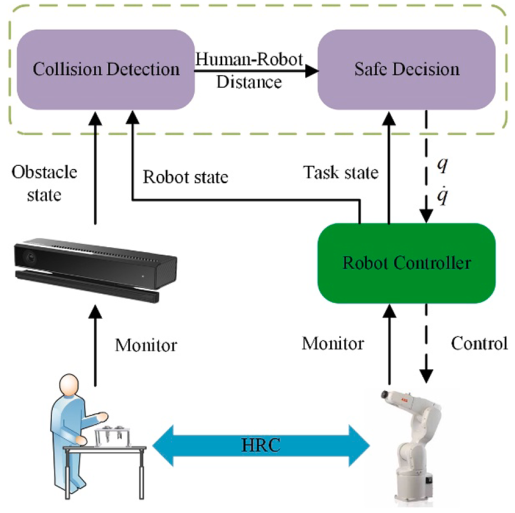}
		\caption{The framework of HRC~\cite{liu2021deep}}
		\label{HRC_architecture}
	\end{minipage}
	\begin{minipage}[t]{0.49\linewidth}
		\centering
		\includegraphics[width=1.9in]{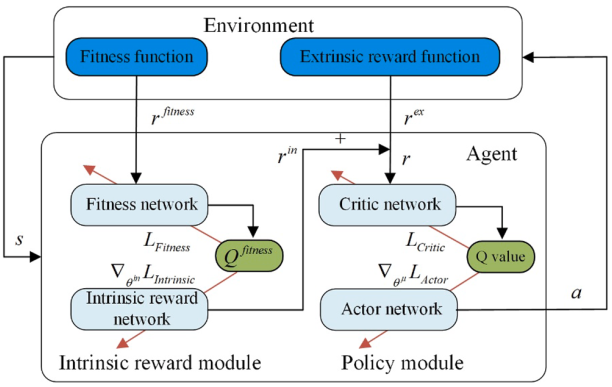}
		\caption{The IRDDPG algorithm~\cite{liu2021deep}}
		\label{HRC_IRDDPG}
	\end{minipage}
\end{figure}

Consider a human working with a robot in a shared space. The HRC problem requires the robot to move to a target position while not colliding with the human, that is, assuring human's safety. As depicted in Figure~\ref{HRC_architecture}, the collision avoidance can be achieved by two models: collision detection and safe decision. The human arm and robot are first modelled by one and four capsules respectively, using capsules bounding boxes\cite{choi2008continuous} (see Figure 3 of \cite{liu2021deep}). Then in the collision detection, the coordinates of joints of the robot are computed by a kinematics model and the skeleton data of the human arm is perceived by Kinect V2~\cite{fankhauser2015kinect}, thereby calculating the distance between the human and robot. Based on the human-robot distance, the robot is controlled to perform safe behaviors to avoid the human arm. DRL has achieved great success in such complex decision-making tasks.

\subsection{Reward Function Design Problem}
Given an appropriate reward, the state-of-the-art DRL algorithms can efficiently maximize the cumulative rewards to solve the HRC problem. Unfortunately, the reward function design could be intractable in HRC, which is therefore the focus of this section.
Due to the multi-objective characteristic of HRC, the reward function as well as the fitness function can be formulated as a weighted sum considering both task completion and risk (show fitness below as an example)
\begin{align}
    R^{\text{fitness}}(s_t, \! a_t)\! \! =\! \!  \lambda_1 R_{\text{risk}}^{\text{fitness}}(s_t,\!  a_t)\!  +\! \lambda_2 R_{\text{task}}^{\text{fitness}}(s_t,\!  a_t),
\end{align}
where $\lambda_1, \lambda_2 \in \Real$ denotes the weight and the fitness $R_{\text{risk}}^{\text{fitness}}(s_t, a_t), R_{\text{task}}^{\text{fitness}}(s_t, a_t)$ are designed in Definition 2 of~\cite{liu2021deep} based on the designer's prior knowledge. Consider a tuple $<S, A, P, R^{\text{fitness}}, T, \gamma, \mathcal{R}>$~\cite{singh2009rewards} that describes the Markov decision process (MDP)~\cite{sutton2018reinforcement} for HRC, which includes a state space $S$, an action space $A$, a transition probability $P$, the fitness, a time horizon $T$, a discount factor $\gamma$ and the reward function space $\mathcal{R}$. The reward design problem then aims to find a reward $R \in \mathcal{R}$ to learn a policy that maximizes the expected cumulative fitness
\begin{align}\label{eqn_reward_design}
    R^\star = \argmax_{R\in \mathcal{R}} \mathbb{E}_{\pi} \left[\sum_{t=0}^{T-1} \gamma^t r_t^{\text{fitness}} | \pi(\cdot|R)\right],
\end{align}
where $\pi(\cdot|R)$ denotes a learned policy given a reward function $R$ and effects of the reward can be passed to the fitness.

\subsection{Intrinsic Reward Deep Deterministic Policy Gradient}
To solve \eqref{eqn_reward_design}, \cite{liu2021deep} proposed an algorithm named intrinsic reward deep deterministic policy gradient (IRDDPG), which is built on the classical DDPG architecture~\cite{lillicrap2015continuous}. Figure~\ref{HRC_IRDDPG} describes the mechanism of IRDDPG that combines an extrinsic reward with a parameterized reward function. IRDDPG consists of a policy module and an intrinsic reward module in which all networks except for the intrinsic reward network have corresponding target networks~\cite{lillicrap2015continuous}, thus 7 networks in total. The fitness network and critic network estimate action value functions $Q^{\text{fitness}}$ on fitness and $Q$ on the total reward respectively, where the latter one relies on the intrinsic reward $r^{\text{in}}$ from the intrinsic reward network and $r^{\text{ex}}$ from the extrinsic reward function. The actor network outputs the action $a$ that is executed by the robot and the environment then feeds back a fitness $r^{\text{fitness}}$ and an extrinsic reward $r^{\text{ex}}$. The training procedure is similar to the classical DDPG using least mean square error (LMS)~\cite{goodfellow2016deep} and policy gradients~\cite{sutton1999policy}. The critic network and fitness network are updated by minimizing the loss function $L_{\text{critic}}$ and $L_{\text{fitness}}$ 
\begin{align}
    &L_{\text{critic}} = \frac{1}{N} \sum_{i=0}^{N-1} (r_i+\gamma Q(s_{i+1}, \mu(s_{i+1})) -Q(s_i, a_i))^2,  \\
    &L_{\text{fitness}} = \frac{1}{N} \sum_{i=0}^{N-1} (r_i^{\text{fit}}+\gamma Q^{\text{fit}}(s_{i+1}, \mu(s_{i+1})) -Q^{\text{fit}}(s_i, a_i))^2. \nonumber
\end{align}
where $\mu$ denotes the deterministic policy. The actor network and intrinsic reward network are updated by policy gradients
\begin{align}
    &\nabla_{\theta^{\mu}} L_{\text{actor}} \approx \frac{1}{N} \sum_{i=0}^{N-1} \nabla_a Q(s_i, a_i)|_{a_i=\mu(s_i)} \nabla_{\theta^{\mu}} \mu(s_i), \nonumber \\
    &\nabla_{\theta^{\text{in}}} L_{\text{intrinsic}} = \nabla_{\theta^{\mu}} J^{\text{fitness}} (\mu) \nabla_{\theta^{\text{in}}} \theta^{\mu}.
\end{align}
where $J^{\text{fitness}}$ denotes the performance function on fitness and $\nabla_{\theta^{\text{in}}} \theta^{\mu}$ is given by (8) in \cite{liu2021deep}. Notice that the policy gradient for the intrinsic reward network is derived by the chain rule.

\subsection{Experiments}
\begin{figure}[!ht]
\centering
\includegraphics[width=3in]{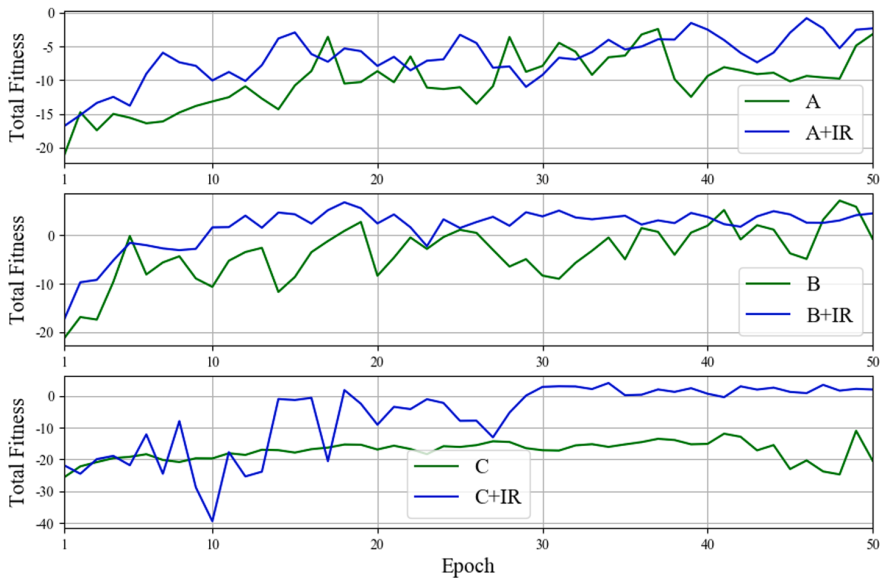}
\caption{Training curves on the total fitness~\cite{liu2021deep}}
\label{HRC_training_curve}
\end{figure}

\begin{figure}[!ht]
	\begin{minipage}[t]{0.49\linewidth}
		\centering
		\includegraphics[width=1.7in]{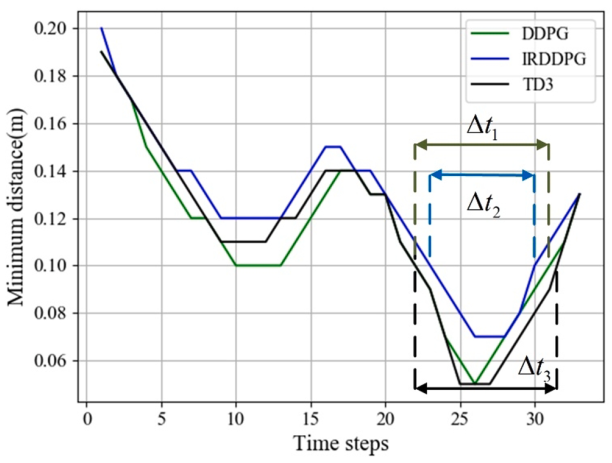}
	\end{minipage}
	\begin{minipage}[t]{0.49\linewidth}
		\centering
		\includegraphics[width=1.7in]{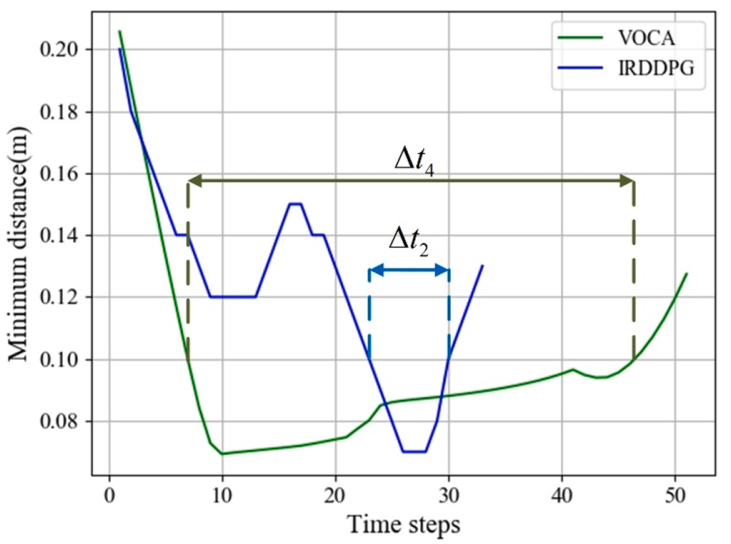}
	\end{minipage}
\caption{The minimum distances of IRDDPG, DDPG, TD3 and VOCA~\cite{liu2021deep}}
\label{HRC_comparison}
\end{figure}

The IRDDPG algorithm is evaluated by several experiments in \cite{liu2021deep} using the ABB industrial robot, where all networks are built by three fully-connected layers. The dimensions of the actor network is (400, 400, 50) and (200, 400, 20) for all other networks. Figure~\ref{HRC_training_curve} depicts the training curves of three groups with different weights contribution (A: $\lambda_1$=1, $\lambda_2$=1; B: $\lambda_1$=2, $\lambda_2$=1; C: $\lambda_1$=1, $\lambda_2$=2). Inside each group there are two experiments with (blue curve) or without (green curve) the parameterized intrinsic reward. Figure~\ref{HRC_training_curve} sees an improved performance of fitness and more stable training curves in either weight contribution by introducing the intrinsic reward.
To compare with other algorithms, Figure~\ref{HRC_comparison} compares the minimal distance of the human and robot using 4 approaches: IRDDPG, DDPG, TD3\cite{fujimoto2018addressing}, VOCA\cite{lin2016human}. As demonstrated, all approaches guarantee the obstacle avoidance at all times and the time of IRDDPG when the minimum distance is less than the safe distance (set to be 0.1 m) is the shortest among algorithms, i.e., $\Delta t_2 < \Delta t_1< \Delta t_3 < \Delta t_4$.
To summarize, the experiments verify the benefits of introducing a parameterized intrinsic reward to a manually designed fitness and extrinsic reward function that are based on one's prior knowledge.
\section{Rapid and Flexible Feature Selection for Deep Reinforcement Learning}
\label{sec_feature_selection}
Given an appropriate reward function, DRL still has issues of data inefficiency and inflexibility, since learning through backpropagation could be prohibitively slow due to the fact that gradients and parameters need to be calculated and updated layer by layer. In addition, the learned representations often tends to be task-specific and the previously learned function will be entirely overwritten when the task changes. To address these issues, one could consider the attention mechanism~\cite{vaswani2017attention} and attend to some useful subsets of representations. This section reviews an approach~\cite{blakeman2022selective} that employs approximate inference instead of backpropagation and combines the selective attention mechanism with particle filters to attend to crucial representations for the current task.
\subsection{The Selective Particle Attention Model}
\begin{figure}[!ht]
\centering
\includegraphics[width=\columnwidth]{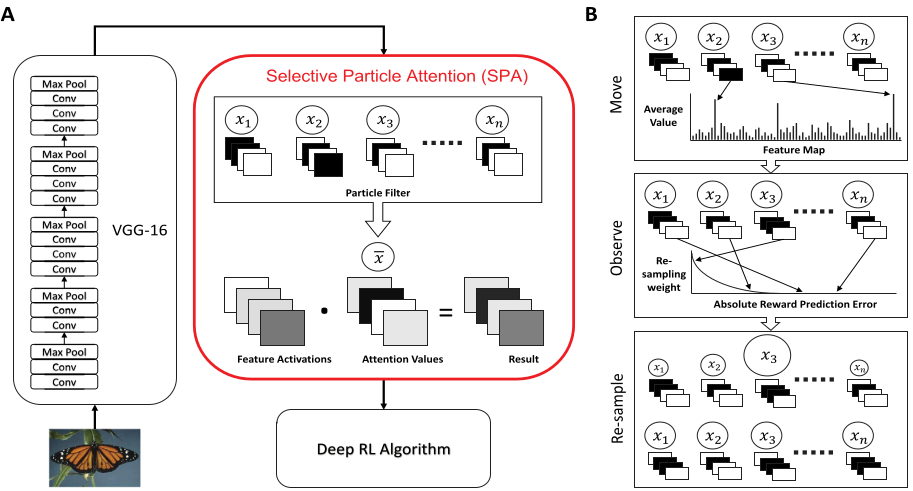}
\caption{The architecture of SPA~\cite{blakeman2022selective}}
\label{spa_architecture}
\end{figure}

As depicted in Figure~\ref{spa_architecture}A, the architecture of the selective particle attention (SPA) proposed by~\cite{blakeman2022selective} consists of three blocks: feature extraction, selective particle attention and DRL algorithms. The first block applies a deep convolutional neural network (CNN)--VGG-16~\cite{simonyan2014very} to process raw pixel inputs to generate the pre-learned representations, since a pre-trained VGG-16 model can be easily leveraged for feature extraction as it has been trained on millions of images~\cite{deng2009imagenet} and are well compatible with raw pixel inputs. The pre-learned feature maps are then passed to SPA which applies particle filters to generate attention values~\cite{vaswani2017attention} that can help identify relevant feature maps for the current task. Eventually, the selected feature maps are fed to DRL algorithms to be mapped to actions by a fully-connected deep neural network (DNN). It is crucial to point out that the SPA algorithm is independent of the first and third blocks. This means that one can select other feature extraction models to replace VGG-16 in the first block and can also consider other state-of-the-art algorithms~\cite{lillicrap2015continuous,schulman2015trust,schulman2017proximal} instead of A3C~\cite{mnih2016asynchronous} that is implemented in \cite{blakeman2022selective}, as long as it can approximate a value function~\cite{sutton2018reinforcement}. 
\subsection{Attention and Particle Filter}
We now focus on the second block--the mechanism of SPA. In the high level, the pre-learned feature maps from VGG-16 will be re-weighted by an attention vector and the most active and useful feature maps are then selected for the current task. The attention vector $\bm{A}$ is $K$-dimensional with each entry belonging to [0, 1], i.e., $\bm{A} \in [0, 1]^K$, where $K=512$ denotes the number of pre-learned features. Subsequently, each entry of $\bm{A}$ is replicated to match the dimension of a feature map and the extended attention is then element-wisely multiplied by the values of each feature map.

On the other hand, obtaining the attention vector relies on particle filters and each of them estimates a latent random variable $X$ that describes a hypothesis about which features are crucial or useful for the current task and thereby the density of the particles can be utilized to approximate the posterior distribution over $X$. Specifically, 1 and 0 are used to denote the useful and useless features respectively and then the latent variable $X$ can be represented by a $K$-dimensional binary vector, i.e., $X = \{0, 1\}^K$. Note that $X$ has $2^K$ possible values, but here one can use only $N$ ($N=250 \ll 2^{512}$) particles to approximate the posterior distribution over $X$. 

Figure~\ref{spa_architecture}B illustrates the detailed process of the particle filter, which contains two main stages: movement and observation. In the movement, let $f_t^k$ denote the mean activation value of feature map $k$ at time $t$. The transition probability that a particle uses for parameters update is given by \cite{blakeman2022selective}
\begin{align}
\label{eqn_particle_update}
  &v_t^k = \frac{f_t^k}{\sum_{j=1}^K f_t^j}, \quad p^k = \frac{\exp(v_t^k \times \tau_{BU})}{\max_j \exp(v_t^j \times \tau_{BU})}, \nonumber \\
    &P(x_{k}' = n) = (p^k)^n (1-p^k)^{1-n}, \, n \in \{0, 1\},
\end{align}
where $v_t^k$ derives from the normalization of $f_t^k$ and then exponentiated and normalized by the maximum value across all feature maps, which ensures that the most active feature will receive a value of 1. Finally, a Bernoulli distribution is used to describe the probability of $k$-th entry of the particle state. The particles are iteratively updated to represent the most active features given the current input, akin to bottom-up attention~\cite{connor2004visual} whose strength is described by $\tau_{BU}$. The higher $\tau_{BU}$ the larger probability of the most active feature will be attended to. 

In the observation, particles are weighted via the likelihood of the return $R_t$ from a given state and the weights are then used to re-sample the particles and update the posterior distribution for the coming time step. The likelihood of the return $P(R_t| x^i)$ and the error $\delta^i$ between the return and predicted state value are computed by regarding normalized particle states as the attention vector, i.e., \cite{blakeman2022selective}
\begin{align}
    &A_k^i = \frac{x_k^i}{\sum_{j=1}^K x_j^i}, \quad \delta^i=(R_t-V(s_t; \bm{A^i}))^2, \nonumber \\
    &P(R_t\mid x^i) \propto \exp(-(\delta^i - \min_j \delta^j) \times \tau_{TD}),
\end{align}
where $i$ and $k$ are for $i$-th particle and $k$-th feature. Note that
the likelihood of the return $R_t$ is proportional to the error $\delta^i$. The observation evaluates the accuracy of each particle’s hypothesis over which representations are crucial to the current task, similar to top-down attention~\cite{connor2004visual}. $\tau_{TD}$ represents the strength of the top-down attention and larger $\tau_{TD}$ imposes stronger penalization of inaccurate hypotheses. Subsequently, the likelihood is normalized to generate a probability distribution that is used to re-sample particles \cite{blakeman2022selective} 
%
\begin{align}
    P(x')=\sum_{i=1}^N P(R_t|x') \mathbbm{1}(x^i=x') / \sum_{i=1}^N P(R_t|x^i).
\end{align}
where $\mathbbm{1}(\cdot)$ denotes the indication function.

After re-sampling, the attention vector is reset to be the normalized mean of $N$ particle states $\Bar{x}_k = 1 /N \sum_{n=1}^N x_k^n, \quad A_k = \Bar{x}_k / \sum_{j=1}^K \Bar{x}_j$ where the normalization is across all feature maps.
\subsection{Experiments}
To verify the effectiveness of the SPA algorithm, \cite{blakeman2022selective} provides two numerical experiments: the multiple choice task and the object collection game. The multiple choice task uses images from Caltech 101 data set~\cite{fei2006one}. At each time, three categories are randomly chosen and one of them is assigned as the ‘target’. Each trial randomly samples one image from three chosen categories, and will yield a reward of +1 if it is from the ‘target’ and 0 otherwise. The target category is changed to another random one after every 50 trials. Therefore, the agent is required to identify reward-related images based on the features of the target category and has to adapt to the changes in reward structure.
\begin{figure}[!ht]
\centering
\includegraphics[width=3in]{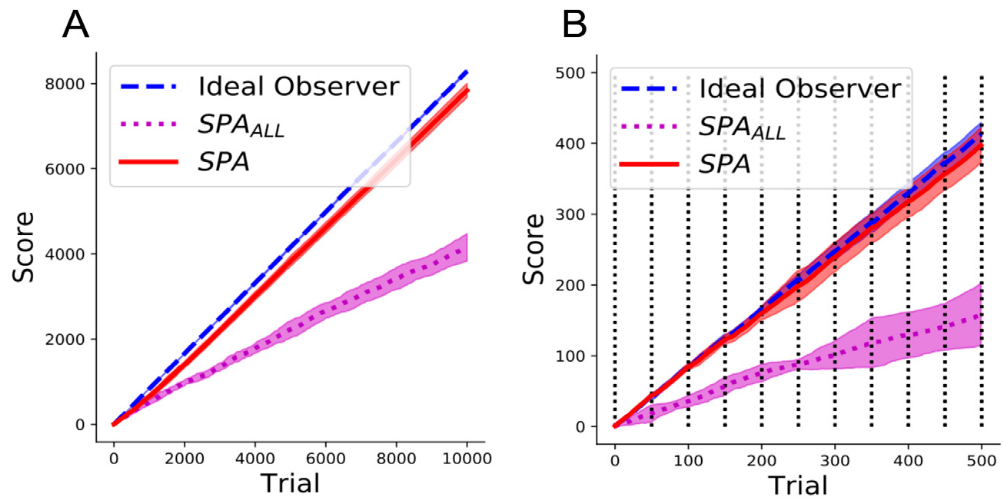}
\caption{Training (A) and test (B) performance of $\text{SPA}, \text{SPA}_{\text{ALL}}$ and Ideal Observer on the multiple choice task~\cite{blakeman2022selective}}
\label{spa_singlestep_decision}
\end{figure}
\begin{figure}[!ht]
\centering
\includegraphics[width=\columnwidth]{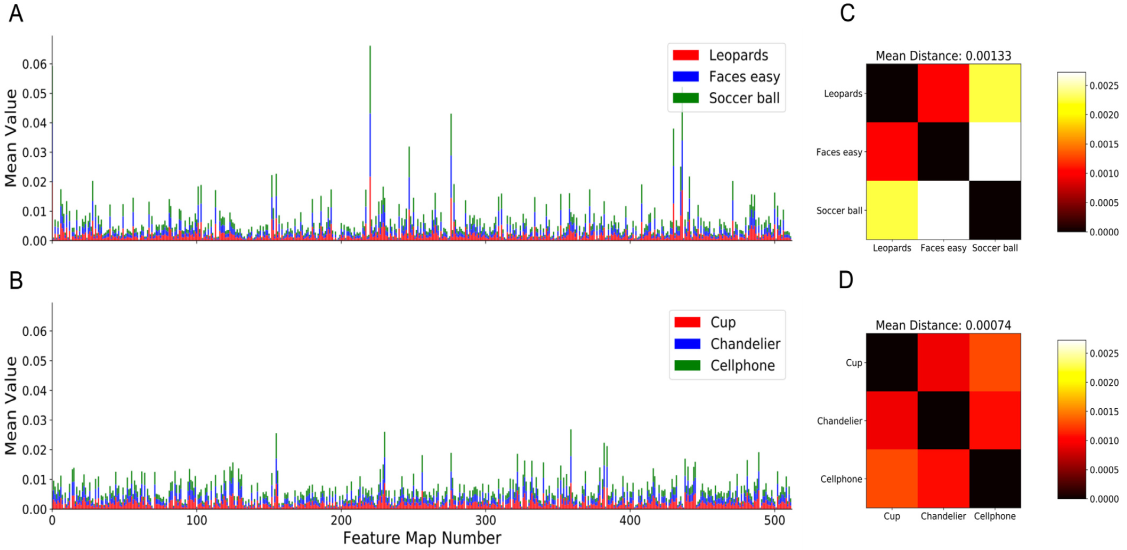}
\caption{Average values of feature maps and corresponding Euclidean distances for the best and worst result~\cite{blakeman2022selective}}
\label{spa_value_feature_map}
\end{figure}

As a comparison, one can consider a naive version $\text{SPA}_{\text{ALL}}$ that assigns each entry of the attention vector to $1/K$, i.e., attending to all features equally. Another model to compare in this task is an ideal observer model that has the ceiling performance, since it chooses the image from the last rewarded category. As shown in Figure~\ref{spa_singlestep_decision}, the SPA model performs close to the ideal observer model on either the training data set (Figure~\ref{spa_singlestep_decision}A) or test data set (Figure~\ref{spa_singlestep_decision}B), while the $\text{SPA}_{\text{ALL}}$ model is significantly worse than the SPA model. The vertical dash line in Figure~\ref{spa_singlestep_decision}B indicates that the reward structure changes with the changing target every 50 trials. Still, SPA shows a great performance that is closed to the ceiling performance. On the other hand, Figure~\ref{spa_value_feature_map}A and B show the mean feature map values from VGG-16 for the image categories that SPA performs best (Leopards, Faces easy and Soccer balls) and worst (Cups, Chandeliers and Cellphones) on. The best case sees several features that are substantially more active than others, while the features take on a more uniform distribution of activation values in the worst case. Figure~\ref{spa_value_feature_map}C and D then illustrate the Euclidean distance between these mean-feature vectors for the best and worst cases, and present that the Euclidean distance of the former is larger than that of the latter.

\begin{figure}[!ht]
\centering
\includegraphics[width=\columnwidth]{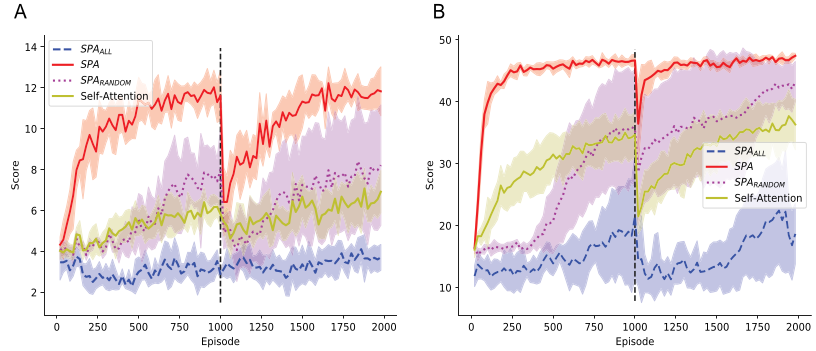}
\caption{Performance of $\text{SPA}, \text{SPA}_{\text{ALL}}, \text{SPA}_{\text{RANDOM}}$ and self-attention on the object collection game~\cite{blakeman2022selective}}
\label{spa_mutistep_decision}
\end{figure}
Figure~\ref{spa_mutistep_decision} describes the object collection game which requires multi-step decision making and is more complicated than the multiple choice task. The details of the setting can be found in~\cite{blakeman2022selective}.
Except $\text{SPA}_{\text{ALL}}$ that is used for comparison before, here another two approaches are implemented to compare with SPA as well. $\text{SPA}_{\text{RANDOM}}$ is similar to $\text{SPA}_{\text{ALL}}$ except that the equal weights is changed to totally random weights. Self-attention follows the classical ``transformer'' model~\cite{vaswani2017attention}. The vertical dash line in Figure~\ref{spa_mutistep_decision}A and B represent that the reward and state space structure are changed at episode 1000. As one can see, SPA outperforms the other three models in all episodes. In addition, after the perturbations at episode 1000, SPA has the largest rate of recovery and shows an earlier onset than the other three. The substantial improvement of performance of SPA highlights again the benefits of approximate inference as opposed to backpropagation. Overall, these experiments verify that SPA is better to deal with the environmental changes and can rapidly and flexibly attend to relevant features for DRL algorithms.

\section{Discussion and Potential Directions}
\label{section_discussion}
\subsection{For IRDDPG Algorithm in HRC}
One of the reasons for DRL being intractable to be applied to the real world is that a good reward is difficult to design. Take HRC as an example, the designer can only manually design the fitness function as well as the extrinsic reward function based on his prior knowledge, while fitness is generally rudimentary and lacks guidance for the agent's actions. In addition, a fixed extrinsic reward function can hardly handle the multi-objective HRC problems that impose a tuning of weight contribution of the objectives. Consequently, a parameterized intrinsic reward function is introduced by \cite{liu2021deep} as a compensation to the extrinsic reward and can be updated iteratively by policy gradients to adapt to the changes of the environment, then yielding the IRDDPG algorithm. However, the IRDDPG is now only evaluated by numerical simulation. Naturally, future work could extend it to real physical robots by enhanced sensing approaches. Additionally, future work could consider the architecture of inverse reinforcement learning~\cite{ng2000algorithms} that learns a reliable and efficient reward function with the assistance of expert behaviors.

In terms of the algorithm itself, the weight contribution in \cite{liu2021deep} is still manually selected, It is thus not clear what the optimal weight contribution is. Future work could consider an adaptive selection of weights using e.g., primal-dual methods~\cite{boyd2004convex}.
Moreover, IRDDPG is compared with the TD3 algorithm in Figure~\ref{HRC_comparison}. Future work could apply the reward design mechanism to the TD3 architecture that improves the stability of DRL training and reduces hyper-parameter tuning compared to DDPG, and therefore yields a more fair comparison with the original TD3 approach.

\subsection{For Feature Selection by SPA}
Learning latent features for DRL can lead to a lower dimensionality of input and thus improves the learning efficiency. In addition, instead of using backpropagation to learn some task-specific representations for DRL, SPA attends to useful subsets of pre-learned features by only reward signals and it allows to rapidly shift attention to new relevant features when the task changes. The object collection game verifies the effectiveness of SPA. However, one may notice that the reward and state space changes are minor due to the fact that the agent only needs to attend to different shape or colors. The task before episode 1000 and after episode 1000 are still similar, by which the useful features for both tasks could overlap. Therefore the benefits of SPA is less convincing. Future work could compare SPA with other state-of-the-art algorithms on two totally different tasks to further verify the improved flexibility of SPA.

A benefit of using particle filters is that they use only 250 particles to approximate the latent variable that has $2^{512}$ possible values. But currently, it is not clear what the optimal number of particles is. Future work could investigate how the number of particles affects the result of feature selection. On the other hand, SPA relies on the nature of the representations that it attends to. For instance, SPA could perform bad if no subsets of features are sufficient for the performance requirements or the underlying representations are not abstract enough~\cite{blakeman2022selective}. Consequently, future work could investigate how the nature of latent representations e.g., disentangled representations~\cite{higgins2018towards} influences the feature selection of SPA.

\section{Conclusions}
\label{sec_conclusions}
In this work, we reviewed two papers that indicate open problems of deep reinforcement learning (DRL) and proposed corresponding solutions. To solve the reward design problem of DRL in human-robot collaboration (HRC), \cite{liu2021deep} proposed an intrinsic reward deep deterministic policy gradient (IRDDPG) algorithm that applies a parameterized intrinsic reward and a manually designed extrinsic reward to the classical DDPG architecture, which improves both performance of completing tasks and guaranteeing human's safety.
\cite{blakeman2022selective} proposed an algorithm named selective particle attention (SPA) built on the attention mechanism where the particle filters employed approximate inference instead of backpropagation. It thereby allows a rapid and flexible update to particles that identify the relevant features for the current task and then improves the efficiency and flexibility of DRL. Some directions for future research have been discussed as well, e.g., combining IRDDPG with the techniques of inverse reinforcement learning and primal-dual methods, and exploring how the nature of latent representations or the number of particles affects the effectiveness of SPA.

\bibliographystyle{ieeetr}
\bibliography{bib}

\end{document}